# Unmanned Aerial Vehicle Instrumentation for Rapid Aerial Photo System

**Widyawardana Adiprawita\***, **Adang Suwandi Ahmad [=] and Jaka Semibiring[+]**

*School of Electric Engineering and Informatics
Institut Teknologi Bandung, Bandung, Id.
e-mail: wadiprawita@stei.itb.ac.id

[=] School of Electric Engineering and Informatics
Institut Teknologi Bandung, Bandung, Id.
e-mail: asa@isrg.itb.ac.id

[+] School of Electric Engineering and Informatics
Institut Teknologi Bandung, Bandung, Id.
e-mail: jaka@ itb.ac.id

**Abstract**

This research will proposed a new kind of relatively low cost autonomous UAV that will enable farmers to make just in time mosaics of aerial photo of their crop. These mosaics of aerial photo should be able to be produced with relatively low cost and within the 24 hours of acquisition constraint. The autonomous UAV will be equipped with payload management system specifically developed for rapid aerial mapping. As mentioned before turn around time is the key factor, so accuracy is not the main focus (not orthorectified aerial mapping). This system will also be equipped with special software to post process the aerial photos to produce the mosaic aerial photo map.

## 1 Introduction

Agriculture is one of the main income sources in Indonesia. Most of the Indonesian citizens have jobs in Agriculture field. Despite this importance of agriculture in Indonesia, there is still lacks of good agriculture practices in Indonesia.

One of the emerging practices in agriculture is "Precision Agriculture". Precision Agriculture refers to the use of an information and technology-based system for field management of crops. Information technology-based system will help the farmer making the right decision. This approach basically means adding the right amount of treatment at the right time and the right location within a field—that's the precision part. Farmers want to know the right amounts of water, chemicals, pesticides, and herbicides they should use as well as precisely where and when to apply them.

By using the tools of precision Agriculture, farmers can specifically target areas of need within their fields and apply just the right amounts of chemicals where and when they are needed, saving both time and money and minimizing their impact on the environment. Irrigation is both difficult and expensive and gets even more difficult when the topography of the terrain is graded. Farmers have a tendency to over irrigate, spending both more time and money than is necessary. Often times farmers look at weather variables and then schedule irrigation based on that information. But if they had better information, they could use scientific models and equations to compute more precisely, how much water their crop is using or how much more is needed. And all this require to have an accurate map of the field. Much of the ability to implement precision agriculture is based on information technologies; in particular, global positioning and navigation and geospatial / remote sensing mapping and analysis.

As mentioned before one of the key technology in precision agriculture is geospatial / remote sensing mapping and analysis. An optimum remote sensing system for precision agriculture would provide data as often as twice per week for irrigation scheduling and once every two weeks for general crop damage detection. The spatial resolution of the data should be as high as 2 to 5 square meters per pixel with positional accuracy of within 2 meters. Additionally, the data must be available to the farmer within 24 hours of acquiring them. Turnaround time is more important to farmers than data accuracy. They would gladly accept remote sensing measurements that are as poor as 75 percent accurate if they were assured of getting them within 24 hours of acquisition. Unfortunately, there are currently no Earth orbiting satellites that can meet all of a precision farmer's requirements. This is where the Autonomous Unmanned Aerial Vehicle (UAV) will play its role.

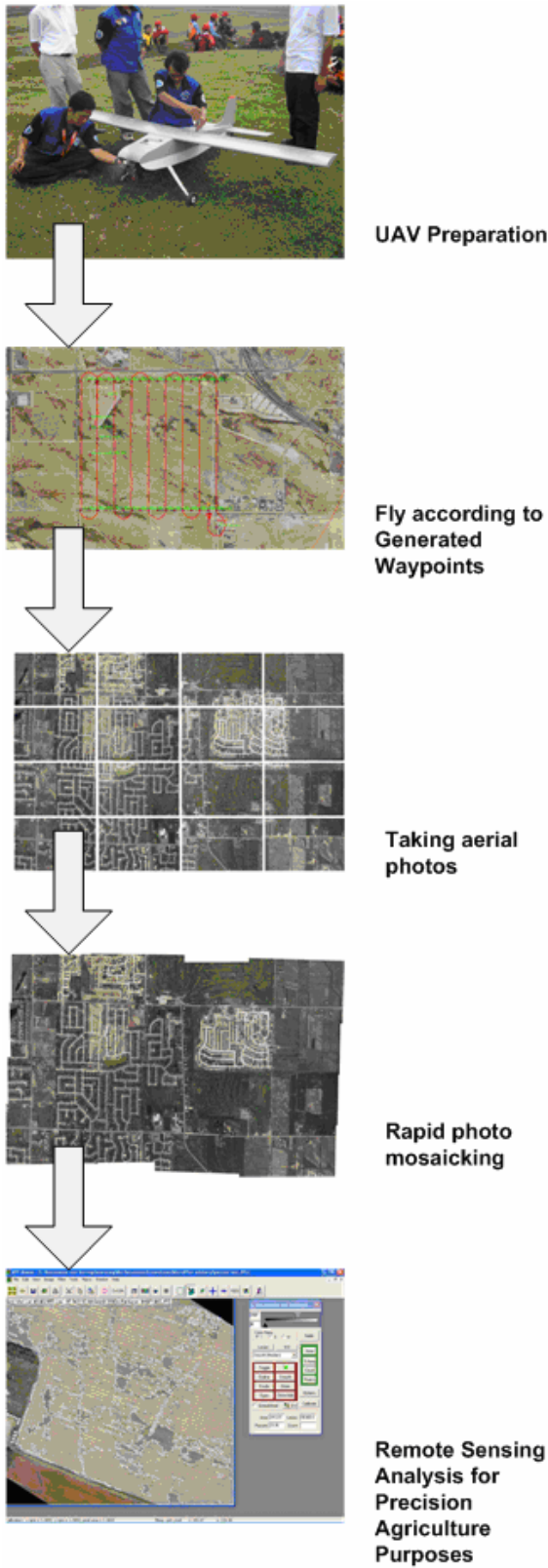

**Figure 1:** The Proposed System

## 2 Research Description

### 2.1 Problem Definition
- Aerial photo of crop fields is needed to enable farmers make the right decision about kind and amount of treatment at the right time and the right location within a field.

- Turn around time of the aerial photo is more important to farmers than data accuracy. Usually the farmers need the information within 24 hours of acquisition.

- A system that enables the farmers to make fast turn around time of the aerial photo of the crop field is needed.

- Cost is important matter, this includes low first time investment and low operational and maintenance cost.

- Ease of operation is also important matter, considering the availability of human resources quality.

### 2.2 Research Objective
- Design and implement an UAV platform which is small enough to be operated from typical crop field in Indonesia without the need of special airstrip. The proposed launching method is by hand, so the UAV platform should be able to do short take off and landing (STOL, short take off and landing).

- Design and implement a low cost autonomous autopilot navigation system that will be used to automatically navigate the UAV to cover the crop field to produce the mosaic aerial photo

- Design and implements a simple flight planning software that will generates waypoints covering the crop field optimally

- Design and implement payload management system onboard the UAV that enable the automatic timing of digital camera shutter release for aerial photo taking

- Design and implement a post processing software that automates the mosaicking process of the aerial photos, so the turn around time will be fulfilled

- For future development : development of specific payload for precision agriculture other than digital camera (such as bio chemical sensor, environmental sensor, weather sensor, etc)

## 3 Methodology

Development of Autonomous Unmanned Aerial Vehicle

1. Airframe : this is the aerial platform that will be instrumented with autonomous autopilot system and carry the mission specific payload (automatic

2. Attitude, Heading and Position Reference System (AHPRS) : this is the main reference input for autonomous autopilot system. The AHPRS outputs euler angles (roll, pitch and yaw), true north absolute heading and position (latitude, longitude and altitude).

3. Autopilot System : this is the main controller of the airframe. It consists of two main part, the low level control system that governs the pose / attitude of the aircraft based on the objective trajectories. The second parts is the waypoint sequencer. This part determined which location the airframe should go (latitude, longitude and altitude), and thus determine the trajectories input the control system.

4. Digital Camera Payload Management System and Automatic Flight Planner : The Flight Planner component will first make automatic waypoint (longitude, altitude, and altitude) that will optimally covers the area of interest to be photographed. Inputs to this systems are boundary of the area (longitude and altitude), scale of the desired aerial photo and horizontal-vertical overlap of each photo segment, then the system will automatically determined the altitude and automatic sequencing of digital camera shutter release. The Digital Camera Payload Management will simply command the shutter release sequence and logging the exact time, oerientation and position of the shutter release (usually recognize as metadata, this information is needed for post processing and automatic rapid mosaicking).

5. Ground Station Software : this component will enable the operator to plan and monitor the mission execution the Autonomous Unmanned Aerial Vehicle, as well as reconfiguring the mission during execution. The monitoring is done in real time because a high speed long range data modem is used to transmit and receive mission parameter between UAV and ground station.

6. Photo indexer software : this software will enable quick view of the relative position of each photo taken during flight. The main objective is to examine the photo coverage and blank spot, so another flight to cover the blank spot can be decided while still on location. Ultimately this software will help the automosaicking process.

7. Final Integration : these steps are taken when all supporting components of the UAV are completely developed.

Development of Post Processing Software

1. Automatic Photo Mosaicking : this component will automatically combined the aerial photographs that covers small area along with the metadata (longitude, latitude and altitude of the digital camera) into single large aerial photographs that covers larger area.

2. Aerial Photo based Agriculture Information System : this is the tools that will be used by the farmers to make analysis to the aerial photograph and support the decision making about the crop field. Remote Sensing and Geographic Information System concepts are involved in this system along with Precision Agriculture good practices.

Overall System Testing : this steps is conducted after the Unmanned Aerial System and Post Processing are completely developed. The objective is to make positive feedback to the overall research and development and to publicize the system to the potential users : the farmers.

Not all steps have been completed on this research. This paper only emphasize the instrumentation side of this research. The airframe for this specific purpose will be developed after the instrumentation have been completed. The Post Processing Software is also in ongoing development.

## 4 Instrumentation of UAV for Rapid Aerial Photo System

### 4.1 Attitude, Heading and Position Reference System (AHPRS)

AHPRS will be used as main reference for autopilot system as well as for Digital Camera Payload Management System.

#### 4.1.1 Attitude Representation

In order to control an aerial platform correctly, one of the input needed by the autopilot control system is attitude. Attitude is usually represented by 3 rotations of the aerial platform. These rotations are roll, pitch and yaw. There are several different rotations representations used. Among them are the euler angle, $C_{bn}$ matrix and the quaternion angle representation. The euler angle (roll $\phi$, pitch $\theta$ and yaw $\psi$) is very intuitive and widely used in aerospace field. But this representation suffers singularity near 90 degree of pitch angle. $C_{bn}$ matrix or direction cosine matrix is a 3x3 matrix that represent sequential rotation of roll, pitch and yaw. This representation doesn't suffer from singularity, but it's not intuitive and uses 9 values to represent attitude. Here is the $C_{bn}$ representation

$$C_{bn}(\phi,\theta,\psi) = \begin{bmatrix} \cos\theta\cos\psi & -\cos\phi\sin\psi + \sin\phi\sin\theta\cos\psi & \sin\phi\sin\psi + \cos\phi\sin\theta\cos\psi \\ \cos\theta\sin\psi & \cos\phi\cos\psi + \cos\phi\sin\theta\sin\psi & -\sin\phi\cos\psi + \cos\phi\sin\theta\sin\psi \\ -\sin\theta & \sin\phi\cos\theta & \cos\phi\cos\theta \end{bmatrix}$$

(1)

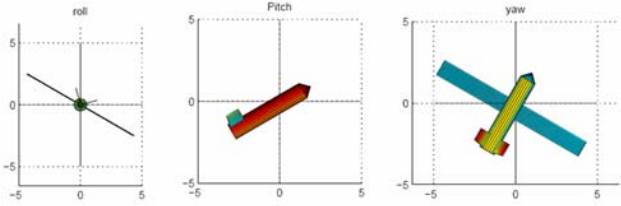

**Figure 2:  Roll, pitch and yaw**

The quaternion representation uses four variables for rotation instead of three. Here is the quaternion representation

$$E = \begin{bmatrix} e0 & e1 & e2 & e3 \end{bmatrix}^T \quad (2)$$

$e0 = \cos(f/2)$

$e1 = \mathbf{A}_x \sin(f/2)$

$e2 = \mathbf{A}_y \sin(f/2)$

$e3 = \mathbf{A}_z \sin(f/2)$

where

A = unit vector along axis of rotation

f = total rotation angle

To measure the attitude there 2 approach that can be used. The first is inertial mechanization which uses discrete time integration over rotation rate measurement. In this approach the attitude determination depends not only to the most current measurement but also depends on the previous attitude value. The second is absolute attitude measurement which can compute the attitude based only on the most current measurement.

### 4.1.2 Body Rotation Rate / Inertial Mechanization

In this approach the attitude is updated using body rotation rate. The body rotation rate is usually represented using a [p q r]$^T$ vector. Each of the value in the body rotation rate vector represents rotation rate in the x, y and z axis in the local coordinate frame of the aerial platform (body frame). The [p q r]$^T$ vector is measured using 3 gyroscopes on each of the x, y and z axis in the body frame. Each Attitude representation has its own formulation for update.

Euler angle change rate formulation :

$$\begin{bmatrix} \dot\phi \\ \dot\theta \\ \dot\psi \end{bmatrix} = \begin{bmatrix} 1 & \sin\phi\tan\theta & \cos\phi\tan\theta \\ 0 & \cos\phi & -\sin\phi \\ 0 & \sin\phi\sec\theta & \cos\phi\sec\theta \end{bmatrix} \begin{bmatrix} p \\ q \\ r \end{bmatrix} \quad (3)$$

$C_{bn}$ change rate formulation :

$$\dot C_{bn} = C_{bn} \begin{bmatrix} 0 & -r & q \\ r & 0 & -p \\ -q & p & 0 \end{bmatrix} \quad (4)$$

Quaternion change rate formulation :

$$\begin{bmatrix} \dot{e0} \\ \dot{e1} \\ \dot{e2} \\ \dot{e3} \end{bmatrix} = \frac{1}{2} \begin{bmatrix} -e1 & -e2 & -e3 \\ e0 & -e3 & -e2 \\ e3 & e0 & -e1 \\ e2 & e1 & e0 \end{bmatrix} \begin{bmatrix} p \\ q \\ r \end{bmatrix} \quad (5)$$

After obtaining the attitude change rate, the result can be integrated in discrete time to obtain the final attitude value.

### 4.1.3 Absolute Attitude Determination

Integration of the Body Rotation Rate (Inertial Rate Mechanization) suffers from the risk of continuous discrete time integration error. So beside the attitude determination that relies on integration and angular rate [p q r]$^T$ measurement, it will be very advantageous if we have attitude determination mechanism that only relies on single measurement in time. Several approaches will be presented, and the sensors involved here are triad of accelerometers (to measure gravity vector) and magnetometers (to measures earth magnetic vector).

#### *4.1.3.1  Roll and Pitch Absolute Determination*

Roll and pitch measurement from gravity vector measured by accelerometer:

$$\theta = -\arcsin(\frac{g_x}{g}) \quad (6)$$
$$\phi = \arcsin(\frac{g_y}{g\cos\theta})$$

And here is the formulation to get the gravity vector :

$$a_{measured} = a_{dynamic} + \omega \times v - g_{body} \quad (7)$$

$a_{measured}$ : 3 axis accelerations measured by accelerometers

$a_{dynamic}$ : 3 axis dynamic accelerations that cause the airframe to move

$\omega \times v$ : centrifugal acceleration, it's the cross product of 3 axis angular rate [p q r]$^T$ and the 3 axis velocity

$g_{body}$ : the gravity vector in body frame

$a_{dynamic}$ and $\omega \times v$ can be obtained from other sensors such as GPS receiver and 3 axis gyroscopes.

#### *4.1.3.2  Yaw Measurement from Earth Magnetic Vector*

Yaw or heading can be calculated from 3 axis earth magnetic field [Mx My Mz]$^T$ and roll - pitch angle using the following equation

Xh = Mx*cos(θ) + My*sin(ϕ)*sin(θ) - Mz*cos(ϕ)*sin(θ)

Yh = Mx*cos(ϕ) + My*sin(ϕ)

ψ = arctan(Yh/Xh)  (8)

It should be noted that this measurement depends on accurate tilt (roll and pitch) measurement, so yaw determination also depends on external reference to get the accurate roll and pitch measurement that free from centrifugal and dynamic acceleration effect.

*4.1.3.3   TRIAD algorithm*

The previous two algorithms treat gravity and magnetic vector field as separate quantities. In TRIAD algorithm we can use both gravity and magnetic vector simultaneously to find the attitude represented by rotation matrix. This problem of determining a rotation matrix needed to rotate one reference vector (a vector reading in inertial / reference frame) to another vector (a vector reading in body frame) is first formulated by Wahba. Several solution has been introduced to solve the Wahba Problem, most of them recursive (i.e. QUEST algorithm). This recursive algorithm to solve Wahba Problem offers the most accurate solution, but from implementation consideration this recursive algorithm will not guarantee the real time aspect needed for the UAV autopilot control system. Fortunately there is one non recursive solution for the Wahba Problem known as TRIAD algorithm. Here is the formulation of the TRIAD algorithm

The gravity vector and magnetic field vector can define a Cartesian coordinate system, with unit vectors along three axes i, j, and k. Two Cartesian systems can be determined by two pairs of ($a_b$, $m_b$) and ($a_R$, $m_R$)

$$\begin{aligned}
i_B &= \frac{a_B}{|a_B|} \\
i_R &= \frac{a_R}{|a_R|} \\
j_B &= \frac{i_B \times m_B}{|i_B \times m_B|} \\
j_R &= \frac{i_R \times m_R}{|i_R \times m_R|} \\
k_B &= i_B \times j_B \\
k_R &= i_R \times j_R \\
C_{bn} &= i_B i_R^T + j_B j_R^T + k_B k_R^T
\end{aligned} \quad (9)$$

**4.1.4   Implementation**

*4.1.4.1   Hardware Implementation*

The Attitude and Heading Reference System will be implemented as an embedded system with the following main components

Microchip's dsPIC30F4013 as the main processor,

- 3 Analog Device's Gyroscope ADXRS150 arranged orthogonal each other to form a gyro triad,
- 1 Freescale's 3 Axis Accelerometer MMA7260Q,
- 1 PNI's Micromag 3 axis magnetometer,
- 1 uBlox TIM LA GPS receiver.

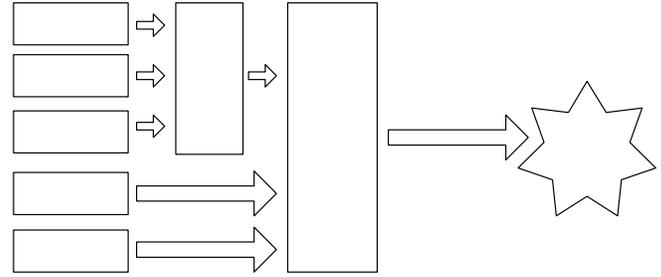

**Figure 3:**   Diagram block of the APHRS Hardware

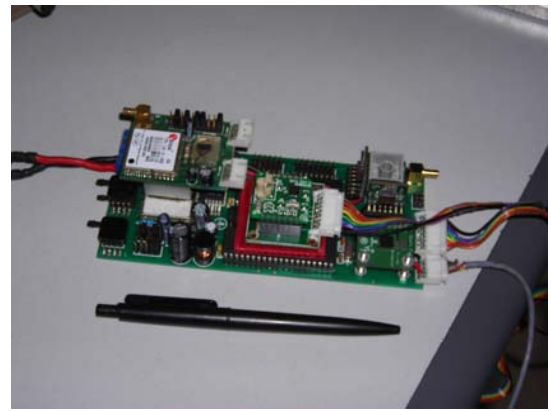

**Figure 4:**   Finished AHRS Hardware

*4.1.4.2   Algorithm and Firmware Implementation*

The firmware is implemented in C language and compiled with Microchip's C30 compiler. The general algorithm is the combination of two algorithm

- rotation rate integration to get high update rate, and
- absolute attitude determination to bound the discrete integration error in rotation rate integration.

The analog sensors (accelerometers and gyros) are sampled in 250Hz and filtered using IIR butterworth 2 order low pass filter. The cut of frequency chosen is 5Hz. This low pass filtering is needed to avoid the vibration caused by UAV's engine. The 5Hz cut of frequency is chosen with the assumption that the UAV's maneuver frequency is below 5Hz. The magnetometer is read through Synchronous Serial Interface / Serial Peripheral Interface (SPI) in 50 Hz update rate. The GPS receiver can give velocity and position updates every 0.25 second or 4Hz.

The Rotation Rate Integration is implemented using quaternion angle representation. The quaternion is updated using gyro reading $[p\ q\ r]^T$. This quaternion angle representation is chosen because it doesn't suffer from singularity problem in 90 degrees pitch. This algorithm is executed in 0.02 second or 50Hz, this frequency is chosen because it's considerably matched to actuator servo frequency.

The absolute attitude determination is implemented using TRIAD algorithm, because this algorithm is considered computationally efficient (no recursive computation) and combines the gravity acceleration and magnetic filled elegantly. To be able to correctly get the gravity acceleration external information of speed and dynamic acceleration from GPS receiver is needed, so this algorithm is executed every 0.25 second or 4 Hz.

So in overall, we have attitude update every 0.02 second (50Hz) and the attitude integration error is zeroed every 0.25 second (4Hz). By combining those 2 algorithms we can have a stable AHRS for considerably long term operation time, including in high G turn (centrifugal or coordinated turn). The fusion algorithm is Kalman Filter.

### 4.1.4.3 Kalman Filter Formulation

Time varying Kalman filter is used to combine the high update rate rotation rate integration (state update) and low update rate absolute attitude determination (measurement update). Here is the AHRS state space formulation

$x_{k+1} = A_k x_k + B_k u_k$ (10)

$z_k = H x_k$

$x = [e0\ e1\ e2\ e3\ bp\ bq\ br]^T$

$u = [p\ q\ r]^T$

$z = [e0\ e1\ e2\ e3]^T$ from TRIAD algorithm solution, with this conversion

$$C_{bn} = \begin{bmatrix} c11 & c12 & c13 \\ c21 & c22 & c23 \\ c31 & c32 & c33 \end{bmatrix}$$

$$e0 = \frac{1}{2}\sqrt{1 + c11 + c22 + c33}$$
$$e1 = \frac{1}{4e0}(c32 - c23)$$
$$e2 = \frac{1}{4e0}(c13 - c31)$$
$$e3 = \frac{1}{4e0}(c21 - c12)$$

(10b)

where

$[e0\ e1\ e2\ e3]^T$ : quaternion attitude representation

$[bp\ bq\ br]^T$ : bias of rotation rate measurements

$[p\ q\ r]^T$ : rotation rate measurements

$$A_k = \begin{bmatrix} 1 & 0 & 0 & 0 & \frac{dt.e1_k}{2} & \frac{dt.e2_k}{2} & \frac{dt.e3_k}{2} \\ 0 & 1 & 0 & 0 & -\frac{dt.e0_k}{2} & \frac{dt.e1_k}{2} & -\frac{dt.e2_k}{2} \\ 0 & 0 & 1 & 0 & -\frac{dt.e3_k}{2} & -\frac{dt.e0_k}{2} & \frac{dt.e1_k}{2} \\ 0 & 0 & 0 & 1 & \frac{dt.e2_k}{2} & -\frac{dt.e1_k}{2} & -\frac{dt.e0_k}{2} \\ 0 & 0 & 0 & 0 & 1 & 0 & 0 \\ 0 & 0 & 0 & 0 & 0 & 1 & 0 \\ 0 & 0 & 0 & 0 & 0 & 0 & 1 \end{bmatrix}$$

$$B_k = \begin{bmatrix} -\frac{dt.e1_k}{2} & -\frac{dt.e2_k}{2} & -\frac{dt.e3_k}{2} \\ \frac{dt.e0_k}{2} & -\frac{dt.e3_k}{2} & \frac{dt.e2_k}{2} \\ -\frac{dt.e3_k}{2} & \frac{dt.e0_k}{2} & -\frac{dt.e1_k}{2} \\ -\frac{dt.e2_k}{2} & \frac{dt.e1_k}{2} & \frac{dt.e0_k}{2} \\ 0 & 0 & 0 \\ 0 & 0 & 0 \\ 0 & 0 & 0 \end{bmatrix}$$

$$H = \begin{bmatrix} 1 & 0 & 0 & 0 & 0 & 0 & 0 \\ 0 & 1 & 0 & 0 & 0 & 0 & 0 \\ 0 & 0 & 1 & 0 & 0 & 0 & 0 \\ 0 & 0 & 0 & 1 & 0 & 0 & 0 \end{bmatrix}$$

Here is the Kalman filter formulation

Time Update (high update rate, 50Hz) :

Project state ahead :

$$\hat{x}_k^- = A\hat{x}_{k-1} + Bu_{k-1}$$ (11)

Project error covariance ahead :

$$P_k^- = AP_{k-1}A^T + Q$$ (12)

Measurement Update (low update rate, 4Hz) :

Compute Kalman Gain :

$$K_k = P_k^- H^T (HP_k^- H^T + R)^{-1}$$ (13)

Update estimate with measurement $z_k$

$$\hat{x}_k = \hat{x}_k^- + K_k(z_k - H\hat{x}_k^-)$$ (14)

Update the error covariance

$$P_k = (I - K_k H)P_k^-$$ (15)

### 4.2 The Autopilot System
The control algorithm of the autopilot system consists of two layers. The first upper layer is waypoint sequencer. The second lower layer is sets of PID (proportional, integrative, and derivative) controller. The waypoint sequencer reads the waypoints given to the autopilot control system by the operator. Each waypoint basically consists of 3D world coordinate which are latitude, longitude and altitude. Based on this waypoint information and current position, attitude and ground speed, the waypoint sequencer will output several objectives: attitude (roll, pitch and yaw/heading objective) and ground speed objectives. These objectives will be read by PID controller as its setting point and will be compared with actual value using PID algorithm to produce servo command value that will actuate the airframe's surface control (aileron, elevator and rudder) and throttle.

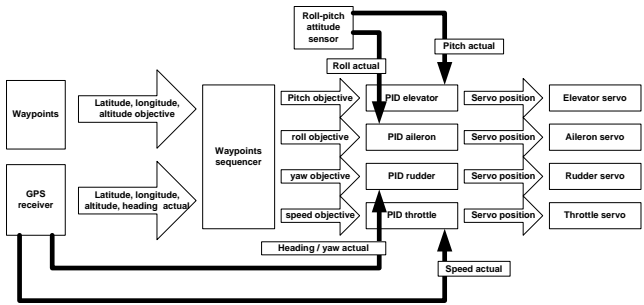

**Figure 5:** Diagram block of the PID Controller

### 4.3 The Flight Planner Software

The flight planner software enable us to define the following requirement :

- area boundary to be photograph (defined as latitude-longitude of left-bottom coordinate and latitude longitude of right-top coordinate)
- the field of view of the camera used
- desired aerial photo scale (will determine the altitude of the trajectory)
- horizontal and vertical photo overlap
- small and large overshoot needed for the airframe to turn direction
- ground altitude
- trajectory direction (north-south or east-west)

and this software will output the following :

- trajectory / waypoints for the autopilot system
- shutter release interval

The waypoints generated by this software have to be uploaded to the autopilot. Here is the screenshot of this software

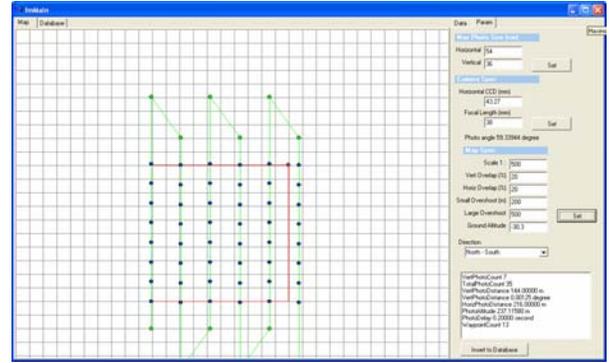

**Figure 6:** The Flight Planner Software

### 4.4 Photo indexer software

As soon as the UAV land, the photo and the metadata containing the exact time, oerientation and position of the shutter release can be downloaded. The photo indexer software will then record the metadata into database and create photo thumbnails automatically. The main purpose of photo indexer is to decide whether the the mission can successfully cover all area or not. If there is any blank spot, this condition can be identified shortly after the UAV land, and another mission can be planned while still on location.

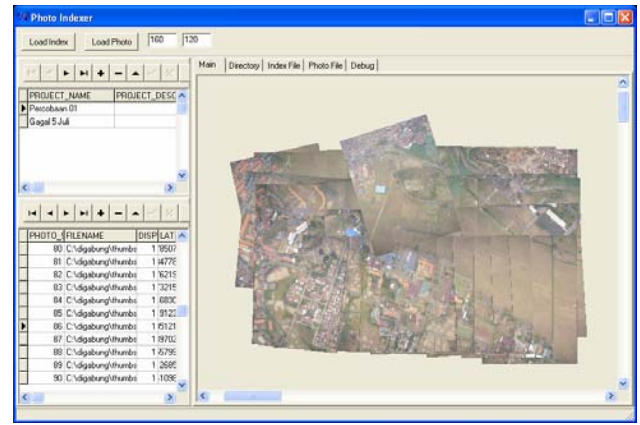

**Figure 7:** Photo indexer screenshot with good coverage

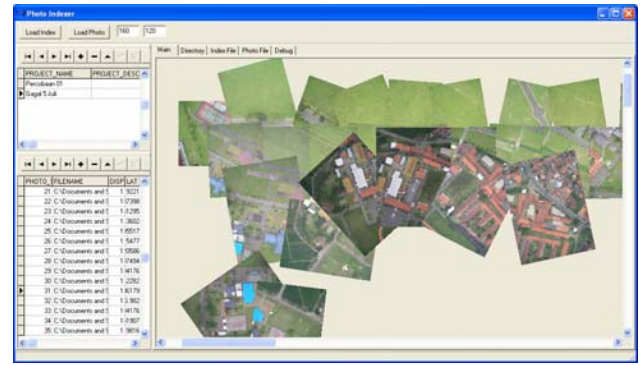

**Figure 8:** Photo indexer screenshot with bad coverage

## 5 Test Result

The system have been tested several time with very good result. Here is one example of the test :

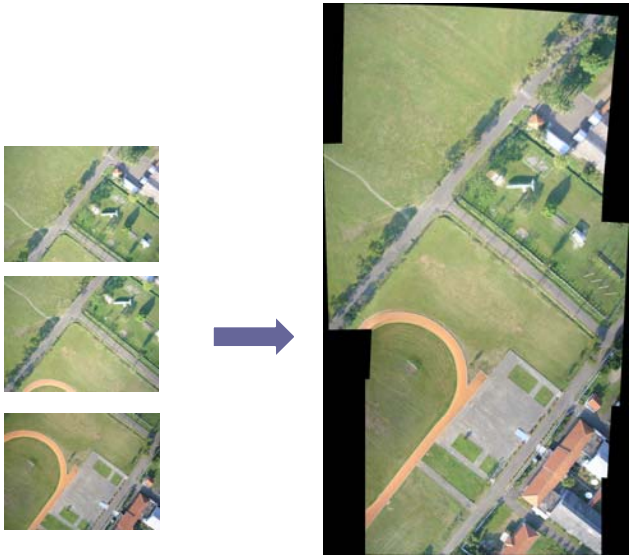

**Figure 9:** Individual Photo and Mosaicking result

## 6 Concluding Remarks

There are still many steps to complete this research. But there are several concluding remarks that can be drawn until now, here they are :

- Mechanical vibration damping for digital camera has to be improved. Further examination need to be conducted to measure the spectrum of the vibration on the camera platform to effectively damp it. Successfully damping the camera platform vibration will contribute directly to the photo quality.

- The airframe used in this test was easily influenced by wind. For future development of the airframe, this condition should be taken into design consideration.

- During test, take off and landing procedure is very critical. The airframe used for test still needed approximately 100 meters long runway. This happened to be quite difficult to find in several test area. So short take off or maybe hand launched airframe will be very useful.

- More accurate recording of time, orientation and position during shutter photo release will enable more accurate photo mosaiking. This concept is known as direct georeferencing. Conventional method of orthorectified photo mosaicking will need ground control point which is difficult and expensive to obtain. Accurate time, orientation and position will enable the photo mosaicking process more straight forward with little or no gound control point. Further concept like Kalman Smoother can be used here.